\documentclass[letterpaper, 10 pt, conference]{ieeeconf}  

\IEEEoverridecommandlockouts                              

\usepackage{multirow}
\usepackage{rotating}
\usepackage{amssymb}
\usepackage{amsmath}
\usepackage[mathscr]{euscript}
\usepackage{tikz}
\def\checkmark{\tikz\fill[scale=0.4](0,.35) -- (.25,0) -- (1,.7) -- (.25,.15) -- cycle;}

\newcommand*{\proposed}{SMAC-Seg}

\title{\LARGE \bf
SMAC-Seg: LiDAR Panoptic Segmentation via Sparse Multi-directional Attention Clustering}

\author{
Enxu Li*,
Ryan Razani*,
Yixuan Xu,
and Liu Bingbing\\
  Huawei Noah's Ark Lab, Toronto, Canada \\
  \texttt{\{thomas.enxu.li, ryan.razani, richard.xu2 liu.bingbing\}@huawei.com} 
  \thanks{ Indicates equal contribution.}
}

\begin{document}

\maketitle
\thispagestyle{empty}
\pagestyle{empty}

\begin{abstract}

Panoptic segmentation aims to address semantic and instance segmentation simultaneously in a unified framework. However, an efficient solution of panoptic segmentation in applications like autonomous driving is still an open research problem. In this work, we propose a novel LiDAR-based panoptic system, called \proposed. We present a learnable sparse multi-directional attention clustering to segment multi-scale foreground instances. \proposed~is a real-time clustering-based approach, which removes the complex proposal network to segment instances. Most existing clustering-based methods use the difference of the predicted and ground truth center offset as the only loss to supervise the instance centroid regression. However, this loss function only considers the centroid of the current object, but its relative position with respect to the neighbouring objects is not considered when learning to cluster. Thus, we propose to use a novel centroid-aware repel loss as an additional term to effectively supervise the network to differentiate each object cluster with its neighbours. Our experimental results show that \proposed~achieves state-of-the-art performance among all real-time deployable networks on both large-scale public SemanticKITTI and nuScenes panoptic segmentation datasets.

\end{abstract}

\section{INTRODUCTION}

Scene understanding is a crucial task in many applications such as autonomous driving and robotics, attracting research attention in domains like computer vision and deep learning. Recently, the topic of panoptic segmentation is introduced to unify the instance segmentation and semantic segmentation in a single trainable network. The purpose of panoptic segmentation is to identify the class labels for points/pixels in the ``stuff" classes and both class labels and instance ID’s for points/pixels in the ``thing" classes. ``Thing" is referred  to  all  countable objects such as pedestrians, cars, and bikes, while ``stuff" is referred to uncountable semantics (background) such as building, sidewalk, and road.

Panoptic segmentation in the image domain  \cite{mohan2021ijcv, Cheng2020panopticdeeplab} has reached a mature state thanks to the structured representation of input images being processed by standard convolutional networks. However, there are few panoptic segmentation methods presented for LiDAR point cloud \cite{9340837, razani2021gps3}. LiDARs have become a pivotal sensor modality in perception applications used in autonomous driving and robotics due to their accurate geometry and light independence. Nonetheless, the sparsity and non-uniform density of point clouds pose new challenges. 
\begin{figure}[htb!]
    \centering
    \includegraphics[width=\linewidth]{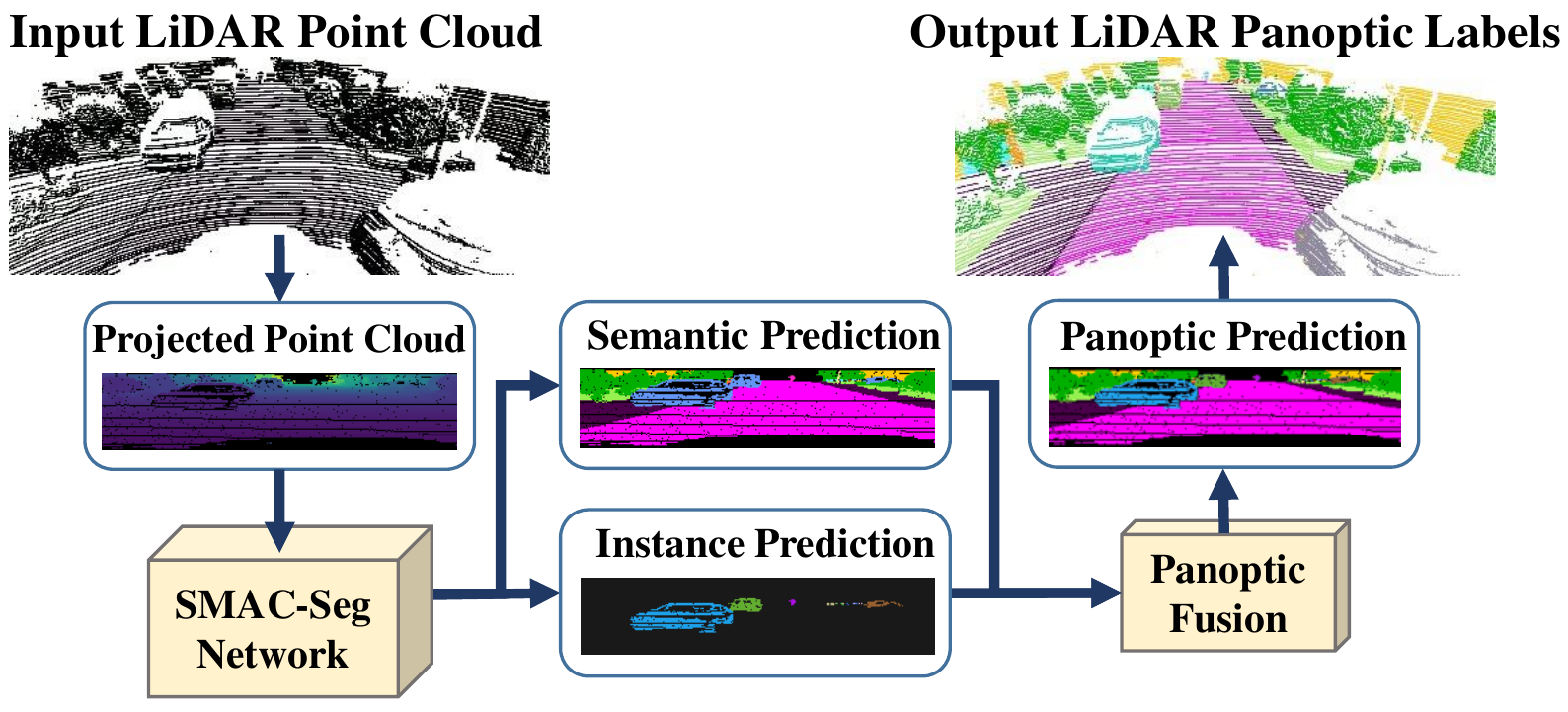}
    \caption{Overview of LiDAR-based \proposed~architecture.}
    \label{fig:overview}
\end{figure}
To this end, several deep learning approaches have been introduced to address these problems. They either process the input point cloud directly \cite{qi2017pointnet, thomas2019kpconv}, divide the 3D scene into cubical or cylindrical voxel grids \cite{cheng2021af2s3net, zhou2020cylinder3d}, or project the point cloud into a 2D image plane either in spherical range-view (RV) \cite{cortinhal2020salsanext, milioto2019rangenet++, razani2021litehdseg} or top-down birds-eye-view (BEV) \cite{zhang2020polarnet,10.1007/978-3-030-11009-3_11}. 
While the first two categories achieve high accuracy compared to projection-based, they are inefficient and require large memory consumption. In contrast, the projection-based can achieve real-time operation but suffers from information loss due to projection operation.
Therefore, there is a need to address the panoptic segmentation problem with a method that offers a real-time and lightweight computation while alleviating the problems due to the projection of point clouds.

In this work, we propose a novel LiDAR-based panoptic system, called \proposed, as depicted in Fig.~\ref{fig:overview}.
We adopt a 2D projection-based approach for the semantic backbone due to its capability of running in real-time. To recover the information lost during projection, we propose a novel convolution with Cross Local Spatial Attention (CLSA). Further, we introduce a learnable clustering module called SMAC to dynamically aggregate points with kernels in multiple directions, relocating points towards their centroids and speeding up the clustering process. Most clustering-based approaches in the literature  \cite{Zhou2021PanopticPolarNet, hong2020lidar} are supervised with L2 loss to regress instance centers by penalizing the difference between predicted and ground truth centers. However, the relative position of the centers with respect to the neighbouring objects is not considered when learning to cluster. Therefore, we introduce a novel centroid-aware repel loss in the training to ensure each object cluster is differentiated from its closest neighbour.

To summarize, our main contributions are as follows: 1) we propose an end-to-end LiDAR-based proposal-free panoptic segmentation network that efficiently clusters the instances, 2) a novel Sparse Multi-directional Attention Clustering (SMAC) to dynamically aggregate each object cluster in birds-eye-view (BEV) and segment in real-time, 3) a unique CLSA block to recover the 3D geometry information lost due to the process of spherical projection, 4) a novel centroid-aware repel loss to effectively reduce the confusion of each object cluster with its neighbouring objects, and 5) a comprehensive analysis on panoptic segmentation performance of our method against existing methods on two public outdoor datasets, SemanticKITTI \cite{DBLP:conf/iccv/BehleyGMQBSG19}, and nuScenes \cite{caesar2020nuscenes}.

\section{RELATED WORK}
Panoptic segmentation task is firstly introduced in images to provide semantic labels and differentiate each object in the pixel-level \cite{panopticMetric}. Most of the panoptic segmentation works in the LiDAR domain are largely built upon advances made in the image domain. Thus, we first review some major and recent works in image panoptic segmentation, followed by approaches introduced on LiDAR point clouds.

\subsection{Image Panoptic Segmentation}

 The first introduced panoptic segmentation baseline from \cite{panopticMetric} combines outputs from PSPNet \cite{zhao2017pspnet} and Mask R-CNN \cite{he2017maskrcnn} with a fusion module. Subsequently, proposal-based methods \cite{li2019attention, mohan2021ijcv} are the most popular approaches among the literature where researchers use a combination of RPN and Mask R-CNN to segment instances. In addition, there are also a handful of proposal-free approaches. For instance, DeeperLab \cite{DBLP:journals/corr/abs-1902-05093} focuses on locating key points and regressing offsets followed by a clustering algorithm to segment objects. Panoptic-DeepLab \cite{Cheng2020panopticdeeplab} builds on DeeperLab and introduces a dual-decoder structure to regress instance object centers and semantic labels, respectively. In contrast, some research works \cite{gao2020ssap, wu2020bidirectional} have also used graph-based approaches to provide panoptic segmentation results on images.

\subsection{LiDAR Panoptic Segmentation}

As a joint task of semantic and instance segmentation, panoptic task usually builds upon a semantic segmentation network. Existing LiDAR semantic segmentation networks could be categorized as 2D-based \cite{milioto2019rangenet++, razani2021litehdseg}, 3D-based \cite{cheng2021s3net, cheng2021af2s3net, zhou2020cylinder3d, tang2020searching}, and point-based \cite{qi2017pointnet, qi2017pointnet++, hu2020randla, thomas2019kpconv}. 2D-based methods usually process the point cloud in a range view (RV) \cite{milioto2019rangenet++, razani2021litehdseg}, birds-eye-view (BEV) \cite{zhang2020polarnet}, or a combination to obtain multi-view \cite{gerdzhev2021tornadonet}. The biggest advantage of this approach is its high efficiency to achieve real-time performance, benefited from the fast inference speed of 2D CNNs. However, the 3D geometry information is lost in the projected 2D representation, which leads to a lack of performance comparing to 3D-based and point-based approaches. In comparison, 3D-based methods usually voxelize the point cloud and use sparse 3D convolutions to process and obtain the segmentation results, which struggle to operate in real-time despite their state-of-the-art performances. Lastly, point-based approaches process the raw point cloud directly, yet they require high computation resources when the point cloud is large-scale, resulting in the slowest inference speed.

Some early LiDAR panoptic works, as an extended task from semantic segmentation task, focus on two-stage systems where the semantic segmentation network is followed by a detection network to generate proposals for objects in the scene. For instance, \cite{DBLP:journals/corr/abs-2003-02371} uses RangeNet++ \cite{milioto2019rangenet++} or KPConv \cite{thomas2019kpconv} to predict the semantics followed by box proposals from PointPillars \cite{Lang_2019_CVPR_pointpillars} to segment objects. PanopticTrackNet \cite{hurtado2020mopt} is another proposal-based approach where the instance segmentation proposal head is adapted from Mask R-CNN. EfficientLPS \cite{sirohi2021efficientlps} brings \cite{mohan2021ijcv} from the image domain and introduce blocks that learns range-aware features targeting LiDAR point clouds.

Many researchers \cite{9340837, gasperini2021panoster, Zhou2021PanopticPolarNet, hong2020lidar} have also explored using clustering to segment the foreground point cloud into objects. The pioneering panoptic work in the LiDAR domain is LPSAD \cite{9340837}. The authors use a shared encoder with two decoders where the first decoder predicts the semantic embedding and the second decoder regresses object centroids for the foreground. Then, a clustering algorithm is applied to segment instances based on predicted semantic embedding and the predicted object centroids in the 3D space. Panoster \cite{gasperini2021panoster} introduces a learnable clustering to assign instance class labels to every point and uses post-processing techniques such as DBSCAN to merge points located close in the 3D space into the same cluster. DS-Net \cite{hong2020lidar} builds upon meanshift clustering and introduces a learnable dynamic shifting module to shift points in 3D towards the object centroids in an iterative manner.

Recently, GP-S3Net \cite{razani2021gps3} introduces a graph-based approach to do instance segmentation where the network embeds each foreground cluster as a graph node and predicts a connection between each pair of nodes to form instances. However, this approach requires a clustering algorithm (e.g., HDBSCAN) to bridge the semantic and instance network and is not real-time.

\section{PROPOSED METHOD}

\subsection{Network Architecture}

\begin{figure*}
    \centering
    \includegraphics[width=\linewidth]{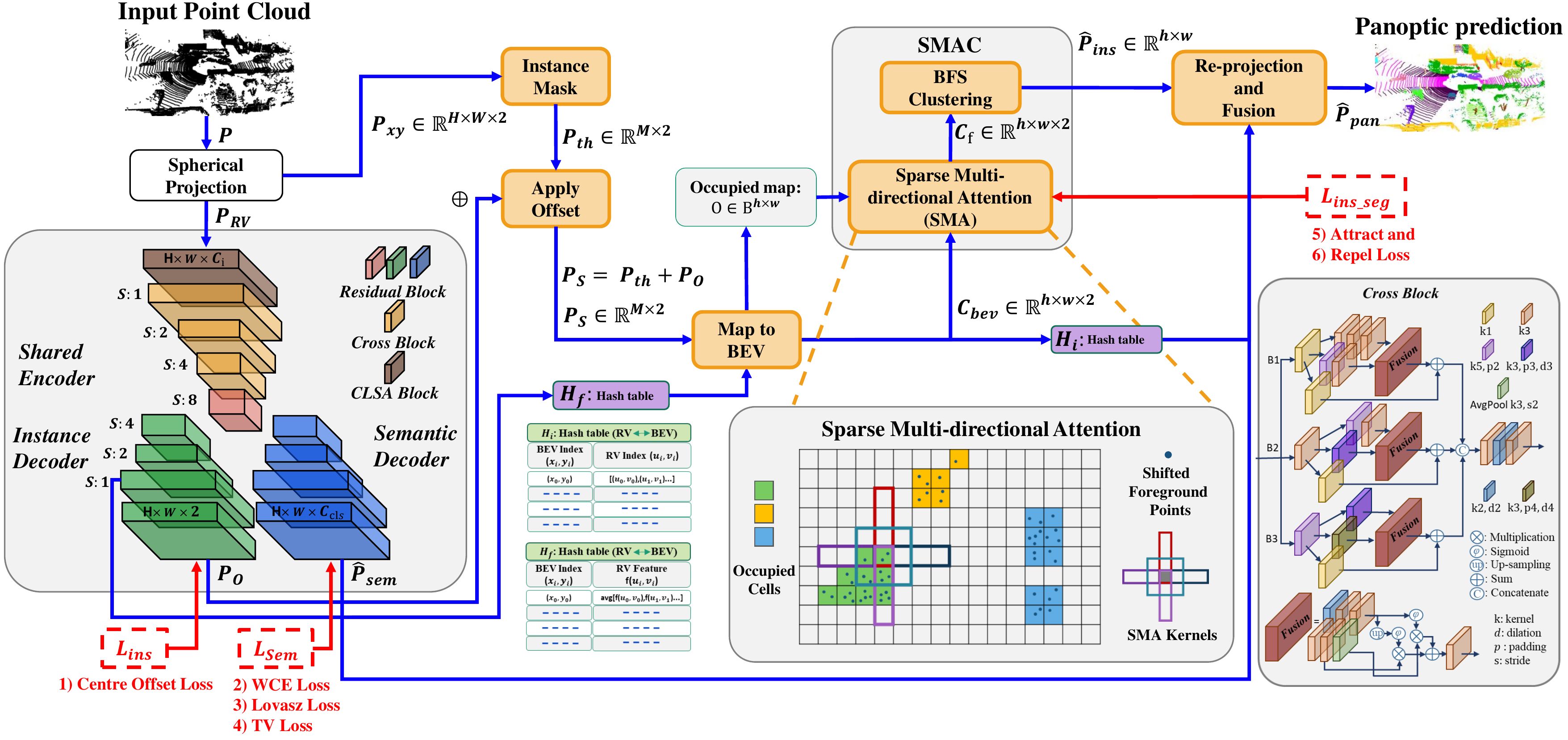}
    \caption{Illustration of SMAC-Seg. The network consists of a shared encoder and dual decoder Unet, and a SMAC module. Please note that the instance mask (top left) is obtained from the ground truth semantic label during training and from the semantic prediction during testing.}%
    \label{fig:SMACSeg}
\end{figure*}

A block diagram of our proposed \proposed~is illustrated in Fig.~\ref{fig:SMACSeg}. The input 3D point cloud is denoted as $\textbf{P} = \{(x,y,z,r, l_{sem}, l_{ins})_{i} \mid i \in \{1,...,N \}  \}$ where $N$ is the number of points in the LiDAR point cloud; $(x,y,z)$ are the Cartesian coordinates in the reference frame centered at the LiDAR sensor; $r$ is the measure of reflectance returned by the LiDAR beam. The input LiDAR point cloud $\textbf{P}$ is first projected into a RV image, denoted as $\mathbf{P}_{RV} \in \mathbb{R}^{H \times W \times C_{i}}$, where $H, W$ are the height and width of the range image and $C_{i}$ is the input features (Cartesian coordinates, remission and depth). The RV image is passed to a shared encoder which includes a CLSA block followed by three Cross blocks and one residual bottleneck block to extract contextual and global features. Then, the down-sampled feature maps are further processed by a dual decoder to be mapped to the same resolution as input, $H \times W$. The semantic decoder predicts semantic classes, denoted as $\hat{\mathbf{P}}_{sem} \in \mathbb{R}^{H \times W \times C_{cls}}$ with $C_{cls}$ number of classes, while the instance decoder regresses the 2D $x,y$ offset $\mathbf{P}_{O}$ $\in$ $\mathbb{R}^{H \times W \times 2}$.

To further obtain instance IDs for the foreground, we first apply an instance mask to filter the point cloud in RV such that only the \textit{thing} points are remained, denoted as $\mathbf{P}_{th}$ $\in$ $\mathbb{R}^{M \times2}$ where $M$ is the number of remaining foreground points and $2$ is the original $x$ and $y$ coordinates. Note that the mask is obtained from the ground truth semantic labels during training and is computed from the predicted semantic labels during testing. Next, we obtain $\mathbf{P}_{s}$ by using $\mathbf{P}_{O}$, the learned 2D center offset of the corresponding foreground points from the instance decoder and shifting them towards the object centers. Further, we project $\mathbf{P}_{s}$ onto a BEV map, $\mathbf{C}_{bev} \in \mathbb{R}^{h\times w\times 2}$ ($h,w$ are the dimension of the BEV map different from the dimension of the range image), using the shifted and discretized $x$ and $y$ coordinates as indices. Consequently, we generate a binary occupancy mask, $\mathbf{O} \in \mathbb{B}^{h \times w}$, to mark the the occupied cells as valid entries. At the same time, we build a hash table, $\mathbf{H}_{f}$, to keep track of the features of the corresponding location in the BEV map with valid entries as well as another hash table, $\mathbf{H}_{i}$, for their original indices on range image. In case of multiple points getting projected to the same BEV location, we take the mean of their features. 

We then apply Sparse Multi-directional Attention (SMA) to aggregate each cluster in $\mathbf{C}_{bev}$ using attention weights obtained from its corresponding features from $\mathbf{H}_{f}$. Subsequently, BFS clustering with a radius of $r$ is used on $\mathbf{C}_{f}$, BEV map generated as the output of SMA, to differentiate each object thus obtain instance label $\hat{\mathbf{P}}_{ins} \in \mathbb{R}^{h \times w}$. Lastly, we map the instance label back to RV using the hash table, $\mathbf{H}_{i}$. The semantic and instance segmentation RV predictions are then mapped to original 3D domain and are concatenated as panoptic predictions. To address any conflicts between semantic and instance predictions, we use majority-voting.

\noindent \textbf{Cross Local Spatial Attention (CLSA)}

\begin{figure}[htb!]
    \centering
    \includegraphics[width=\linewidth]{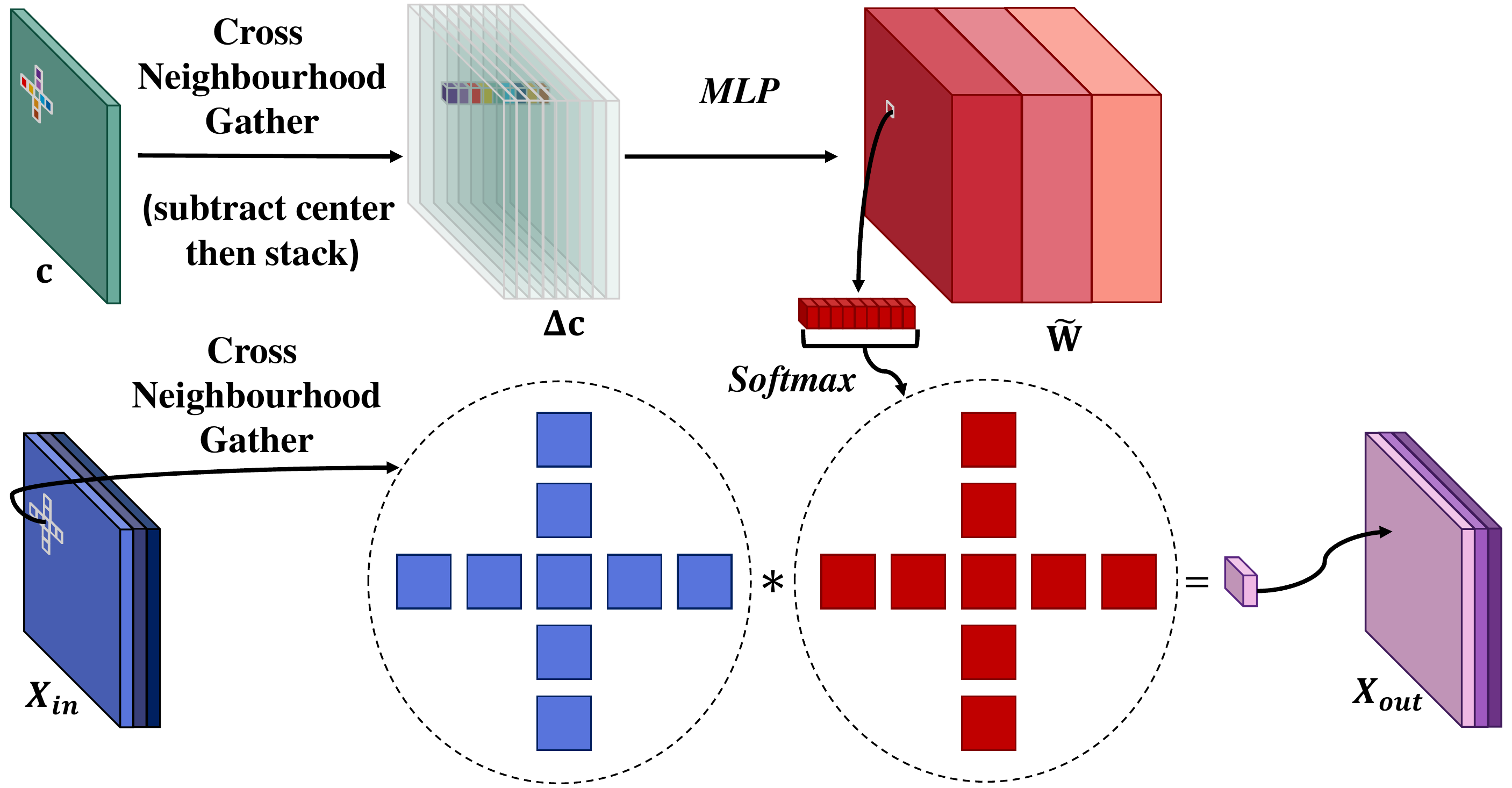}
    \caption{Convolution with Cross Local Spatial Attention (CLSA).}
    \label{fig:CLSA}
\end{figure}

We extend the Diamond Feature Extractor module in \cite{gerdzhev2021tornadonet} and introduce a spatially adaptive feature extractor for RV images to incorporate the local 3D geometry as shown in Fig. \ref{fig:CLSA}. Specifically, we replace the regular convolutions in the second half of the Diamond Block with CLSA convolutions. A 2D convolution operation can be written as,
\begin{equation}
    \mathbf{x}^{out}_{\mathbf{u}} = [\mathbf{x}^{in}_{\mathbf{u}} * \mathbf{W}]_{\mathcal{V}^{2}(K)} = \sum_{\mathbf{i} \in \mathcal{V}^{2}(K)} \mathbf{W}_{\mathbf{i}} \mathbf{x}^{in}_{\mathbf{u}+\mathbf{i}}
\end{equation}
where $\mathbf{u}$ denotes the 2D index to locate each point in the feature map; $\textbf{W} \in \mathbb{R}^{K \times K \times N^{out} \times N^{in}}$ is the kernel weight, shared among each sliding window, with $N^{in},N^{out}$ being the number of input and output feature channels respectively; $\mathcal{V}^{2}(K)$ is the list of offsets in 2D square with length $K$ centered at the origin. Here, we would like $\textbf{W}$ to be adaptive to the geometry of each neighbourhood, in particular, with attention built from the relative positions of the points. Formally, we introduce a 2D convolution with cross local spatial attention as the following:
\begin{equation}
    \mathbf{x}^{out}_{\mathbf{u}} = [\mathbf{x}^{in}_{\mathbf{u}} * \Tilde{\mathbf{W}}_{\mathbf{u}}]_{\mathcal{N}^{2}(K)}
\end{equation}
\begin{equation}
    \Tilde{\mathbf{W}}_{\mathbf{u}} = \sigma \Big[w \Big( \underbrace{\bigcup_{\mathbf{i} \in \mathcal{N}^{2}(K)} \mathbf{c}_{\mathbf{u}+\mathbf{i}} - \mathbf{c}_{\mathbf{u}}}_{\Delta \mathbf{c}_{\mathbf{u}}} \Big)\Big]
\end{equation}
where $\Tilde{\textbf{W}} \in \mathbb{R}^{(2K-1) \times N^{out} \times N^{in} \times H \times W}$ is the spatially adaptive kernel weight computed from the relative geometric positions of the points within the \textit{cross-shaped} neighbourhood; $w(.)$ is a 3-layer MLP; $\mathbf{c}_{\mathbf{u}}$ is the corresponding spatial coordinate feature of $\mathbf{x}_{\mathbf{u}}^{in}$ (i.e. Cartesian $xyz$, depth, and occupancy); $\bigcup$ is the concatenation operator; $\sigma$ denotes the softmax operation on the spatial dimension to ensure the attention weights in the neighbourhood for each feature channel are summed up to 1; and $\mathcal{N}^{2}(K)$ is a set of offsets that define the shape of a cross kernel with size of $K$. e.g. $\mathcal{N}^{2}(3)$ $=$ $\{(-1,0),$ $(0,0),$ $(1,0),(0,1),(0,-1)\}$. We compare the cross kernel sampling with other regularly used kernel shapes (e.g. square kernel) in section \ref{sec:ablation}. \\

\noindent \textbf{CrossUnet}
Our proposed CrossNet backbone is depicted at the bottom right of Fig.~\ref{fig:SMACSeg}. It includes 3 layers of Cross blocks which are designed to capture multi-scale features, followed by a bilateral fusion to obtain rich information at each block. In particular, the input feature is first passed to a multi branch convolution layers and each branch further processes the features with Convolution layers of different receptive field (followed by Relu and BatchNorm layer) to obtain fine-grained information. Next, a Bilateral fusion module is applied on each branch to fuse the features of different resolutions. Finally, all the feature maps are concatenated and their channel numbers are reduced through a final convolution layer for efficient processing.\\

\noindent \textbf{Sparse Multi-directional Attention (SMA)} 
The foreground point cloud in BEV, $\mathbf{C}_{bev}$, is processed by SMA to ensure the points are more aggregated towards the object centers; thus, a simple and fast clustering algorithm like BFS can easily differentiate each cluster.
Further, for every valid entry in $\mathbf{C}_{bev}$, we obtain the center of mass of its neighbourhood in five directions using Eq.\ref{eq:com}, denoted as $\mathbf{C}_{W},\mathbf{C}_{E},\mathbf{C}_{N},\mathbf{C}_{S}$, and $\mathbf{C}_{C}$.

\begin{equation}
    \label{eq:com}
    \mathbf{C}_{\mathbf{u}} = \frac{1}{n_{\mathbf{u}}} \sum_{\mathbf{i} \in \Omega(K)}^{}\mathbf{C}_{bev, \mathbf{u+i}} \cdot \mathbf{O}_{\mathbf{u+i}}
\end{equation}

\noindent where $\Omega(K)$ is a set of 2D indices for each neighbourhood sampling region with size $K$ and $n$ is the number of valid entries in the neighbourhood acting as a normalization factor, formally, $n_{\mathbf{u}} = \sum_{\mathbf{i} \in \Omega}^{} \mathbf{O}_{\mathbf{u+i}}$. We denote multi-directional neighbour sampling as $\Omega_{W}(K):\{(0, i)$  $\mid$   $\forall i \in [-K, 0]\}$, $\Omega_{E}(K):\{(0, i)$  $\mid$   $\forall i \in [0,K]\}$, $\Omega_{N}(K):\{(i, 0)$  $\mid$   $\forall i \in [-K,0]\}$, $\Omega_{S}(K):\{(i, 0)$  $\mid$   $\forall i \in [0,K]\}$. The foreground point cloud at location $\mathbf{u}$ in BEV representation after being processed by  SMA module, $\mathbf{C}_{f, \mathbf{u}}$, can be expressed as, 
\begin{equation}
    \mathbf{C}_{f, \mathbf{u}} = \mathbf{C}_{all, \mathbf{u}} \times \underbrace{\sigma(MLP(\mathbf{H}_{f, \mathbf{u}}))}_{\mathbf{\pi}_{\mathbf{u}}}
\end{equation}
where $\mathbf{C}_{all, \mathbf{u}} \in \mathbb{R}^{2\times5} = cat( \mathbf{C}_{W,\mathbf{u}}$,$\mathbf{C}_{E, \mathbf{u}}$,$\mathbf{C}_{N, \mathbf{u}}$,$\mathbf{C}_{S, \mathbf{u}}$,$\mathbf{C}_{C, \mathbf{u}})$ is the concatenated $xy$ centers of mass from applying kernels in five directions at location $\mathbf{u}$; $\mathbf{\pi}_{\mathbf{u}} \in \mathbb{R}^{5}$ is the attention weights computed using the MLP from the foreground features at location $\mathbf{u}$ of the BEV map; $\sigma$ denotes the softmax operator to ensure the attention weights in all directions summing up to 1, and $\times$ is matrix multiplication. Essentially, $\mathbf{C}_{f}$ is the final location of the foreground point cloud in BEV, shifted towards its neighbouring points after receiving the directional guidance from the network.

\subsection{Centroid-aware Repel Loss}
The main purpose of the SMAC module is not to have an accurate prediction of the object center, but to have each object forming a cluster that could be easily differentiated from others in the 2D BEV space. In order to tackle this problem, we propose a novel Centroid-aware Repel Loss to supervise this module.

\begin{equation}
    \label{eq:repel}
    L_{repel} = \frac{1}{I} \sum_{i=1}^{I} \frac{1}{P_{i}} \sum_{p=1}^{P_{i}} max\{
    0, d_{i} - \hat{d}_{i,p}
    \}
\end{equation}
\begin{equation}
    d_{i} = \min_{j \in [1,I] \cap j \neq i} \lVert \mathbf{C}_{gt,i} - \mathbf{C}_{gt,j}\rVert_{2}
\end{equation}
\begin{equation}
    \hat{d}_{i,p} = \min_{\substack{q \in [1,P_{j}] \\ j \in [1,I] \cap j \neq i}} \lVert \mathbf{C}_{f,(i,p)} - \mathbf{C}_{f,(j,q)}\rVert_{2}
\end{equation}
where $I$ is the total number of instances, $P_{i}$ is the number of occupied points in $\mathbf{C}_{bev}$ for instance $i$, $\mathbf{C}_{f,(i,p)} \in \mathbb{R}^{2}$ is the final 2D position after $SMAC$ at point $p$ that belongs to instance $i$, and $\mathbf{C}_{gt,i} \in \mathbb{R}^{2}$ is the ground truth 2D centroid of instance $i$. Essentially, $\hat{d}$ represents the closest distance from itself (final shifted position) to any other point from other objects, and $d$ represents the distance between the ground truth centroid of current object to the other closest instance. This loss term penalizes if the ground truth distance, $d$ is larger than $\hat{d}$, meaning the network still needs to learn such that each foreground cluster is repelled from others.

\begin{figure}[htb!]
    \centering
    \scalebox{0.95}
    {
    \includegraphics[width=\linewidth]{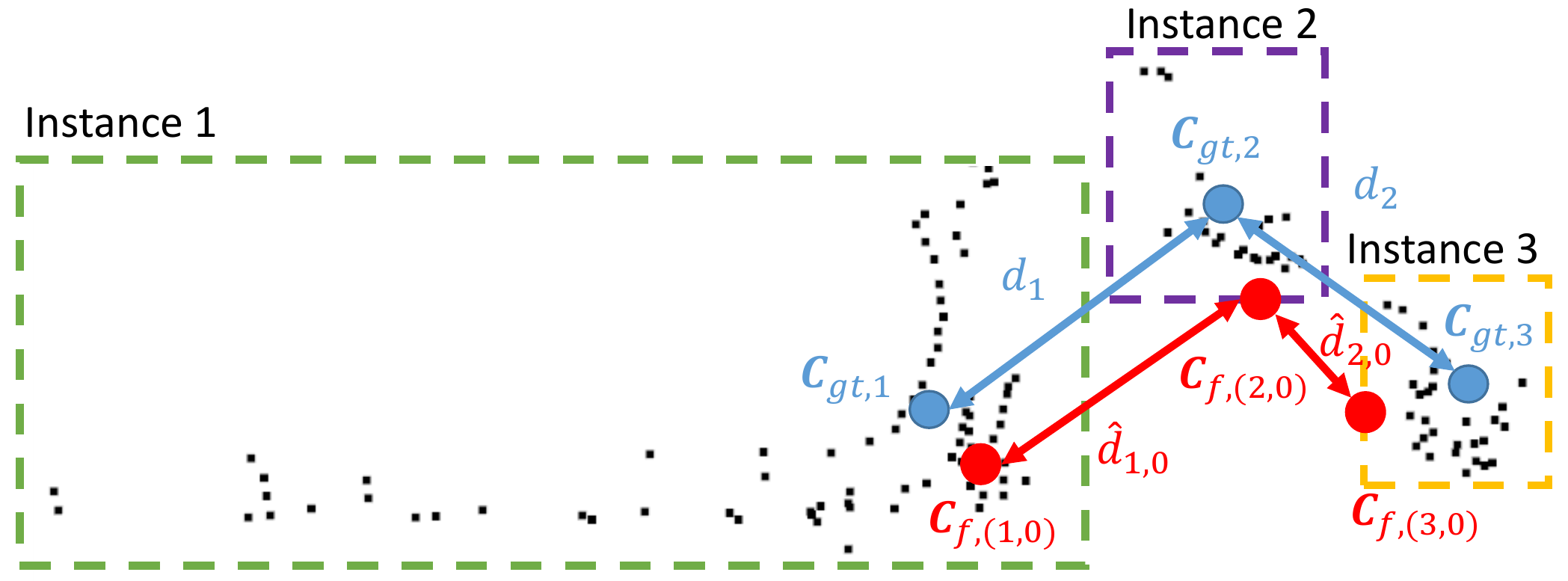}
    }
    \caption{Illustrations of how the centroid-aware repel loss penalizes the network. Three instances are present in the scene where black points are from the original point cloud in BEV, cyan points are the GT centroid for each instance, and red points are the final location of the points after SMAC. In this case, the final predicted centroid of instance 2 and 3 are close (\textcolor{red}{$\hat{d}_{2,0}$} is small compare to ground truth \textcolor{cyan}{$d_{2}$}). This loss function will repel $\mathbf{C}_{f,(2,0)}$ away from $\mathbf{C}_{f,(3,0)}$.
    }
    \label{fig:repel}
\end{figure}
Further, we use an additional loss term to enforce the variance of each cluster is minimized, as used in \cite{9340837}. 
\begin{equation}
    L_{attract} = \frac{1}{I} \sum_{i=1}^{I} \frac{1}{P_{i}} \sum_{p=1}^{P_{i}} 
    \lVert \mathbf{C}_{f,(i,p)} - \Bar{\mathbf{C}}_{f, i}\rVert_{2}
\end{equation}
where $\Bar{\mathbf{C}}_{f, i}$ is the average of the all the point locations in BEV after SMAC for instance $i$ and the rest of the terms are defined as the same as in Eq.\ref{eq:repel}.
We use three loss terms to supervise the semantic segmentation, similar to \cite{gerdzhev2021tornadonet} and L2 regression loss, similar to \cite{9340837,hong2020lidar} to supervise the center offset from the instance decoder. Thus, the total loss is the weighted combination illustrated as follows:
\begin{equation}
    \substack{L_{total} = \underbrace{\beta_{wce}L_{wce} + \beta_{ls}L_{ls} + \beta_{tv}L_{tv}}_{\text{semantic decoder}} + \underbrace{\beta_{l2}L_{l2}}_{\text{instance decoder}} \\
    + \underbrace{\beta_{repel}L_{repel} + \beta_{attract}L_{attract}}_{\text{SMAC}}}
\end{equation}
\section{EXPERIMENTS}

In this section, we discuss the experimental setup and demonstrate the performance of \proposed~on popular driving-scene datasets for panoptic segmentation using 3D LiDAR point clouds. We compared our results with state-of-the-art approaches and conducted ablation studies to evaluate the importance of the main components.

\subsection{Datsets and Evaluation Metrics}

In terms of datasets, we used \textbf{SemanticKITTI} dataset \cite{DBLP:conf/iccv/BehleyGMQBSG19} and nuScenes dataset  \cite{caesar2020nuscenes}. SemanticKITTI is the first publically available dataset on LiDAR-based panoptic segmentation for driving scenes. For the panoptic segmentation task, each point in the dataset is assigned a label of 19 classes. The instance IDs are available for humans and vehicles, which account for 8 \textit{things} classes. Other objects without instance IDs belong to \textit{stuff} classes.

\textbf{NuScenes} is a popular large-scale driving-scene dataset, where official point-level panoptic segmentation labels for LiDAR scans are unavailable \footnote{at the time of writing}. Given the 3D bounding box annotations and the semantic labels, we generated our own labels for training and evaluation. Specifically, we assigned the same instance ID for all points that have the same semantic labels and are located within the same bounding box. The 16 labeled classes in the \textit{lidarseg} dataset were divided into 8  \textit{things} classes and 8 \textit{stuff} classes. During evaluation, we followed \cite{Zhou2021PanopticPolarNet} to discard instances that contain fewer than 20 points. Given the official training-validation split, we trained our model on 700 training scenes and provided validation results on 150 validation scenes.
\begin{figure}[htb!]
    \centering
    \includegraphics[width=\linewidth]{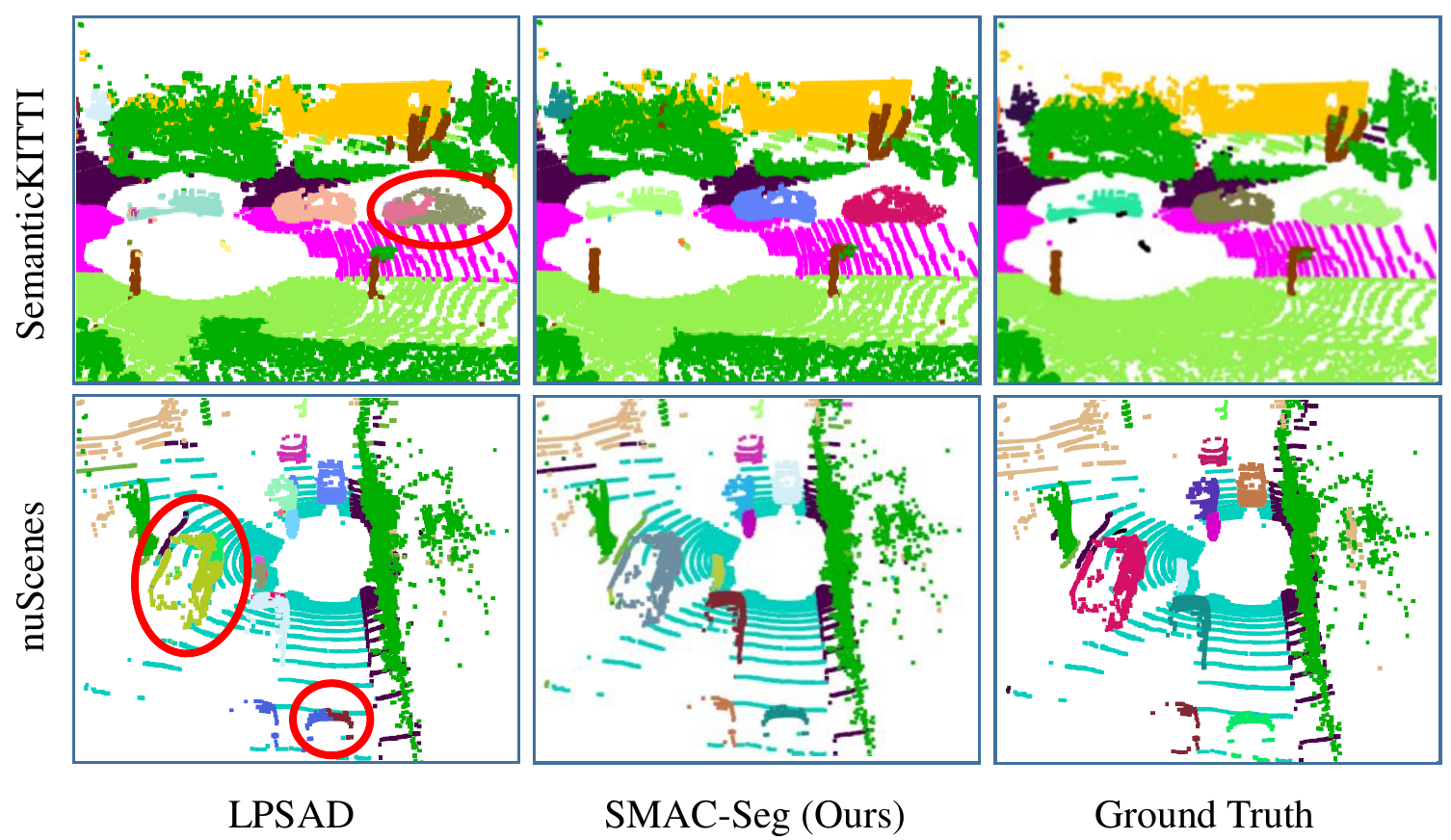}
    \caption{Comparison of SMAC-Seg with LPSAD on both SemanticKITTI and nuScene dataset.}
    \label{fig:qualitative}
\end{figure}
For metrics, we follow \cite{panopticMetric} to use the mean Panoptic Quality (PQ) for evaluation and comparison with other models. We also reported the mean Recognition Quality (RQ) and the mean Segmentation Quality (SQ) and calculated these metrics separately on \textit{stuff} and \textit{things} classes, providing PQ\textsuperscript{St}, SQ\textsuperscript{St}, RQ\textsuperscript{St} and PQ\textsuperscript{Th}, SQ\textsuperscript{Th}, RQ\textsuperscript{Th}.

\subsection{Experimental Setup}
We trained our network end-to-end for 100 epochs on SemanticKITTI training split and evaluated on the test split. All ablation experiments were validated on seq8 (validation set). The range image resolution was set as $64 \times 2048$. For nuScenes, we trained our network for 150 epochs on the training scenes and evaluated on the validation scenes. We used $32 \times 1024$ as the resolution for the range image as the LiDAR sensor has 32 beams. We provided result of an additional hi-res model specifically for nuScenes, incorporating a range image with size $64 \times 2048$ to reduce the information loss. For both datasets, we trained our model using SGD optimizer with a learning rate of 0.01 and weight decay of $10^{-5}$. 4 NVIDIA V100 GPUs were used and the batch size per GPU was 4. The BEV grid size was set to be $0.5$m. The SMAC kernel size was set as $7$. The weights on the loss functions,  $\beta_{wce},\beta_{ls},\beta_{tv},\beta_{l2},\beta_{repel},\beta_{attract}$ were set to be 1.0, 1.0, 5.0, 0.1, 0.1, 0.1 respectively.

We use a dual-decoder U-net based on residual blocks as the baseline where the first decoder predicts the 2D offsets to object centers and second decoder predicts semantic class labels. Then we add a clustering algorithm after the first decoder to segment the foreground into objects. We include both traditional (e.g. BFS, HDBSCAN \cite{campello_hdbscan}, and meanshift \cite{comaniciu_meanshift}) and learning-based (e.g. Dynamic Shifting \cite{hong2020lidar}) clustering in comparison. In particular, the baseline with Dynamic Shifting clustering is adapted from \cite{hong2020lidar} where we attach the DS clustering head to our first decoder.
\subsection{Quantitative Evaluation}
\label{sec:quanti}
\begin{figure}[htb!]
    \centering
    \scalebox{0.95}
    {
    \includegraphics[width=\linewidth]{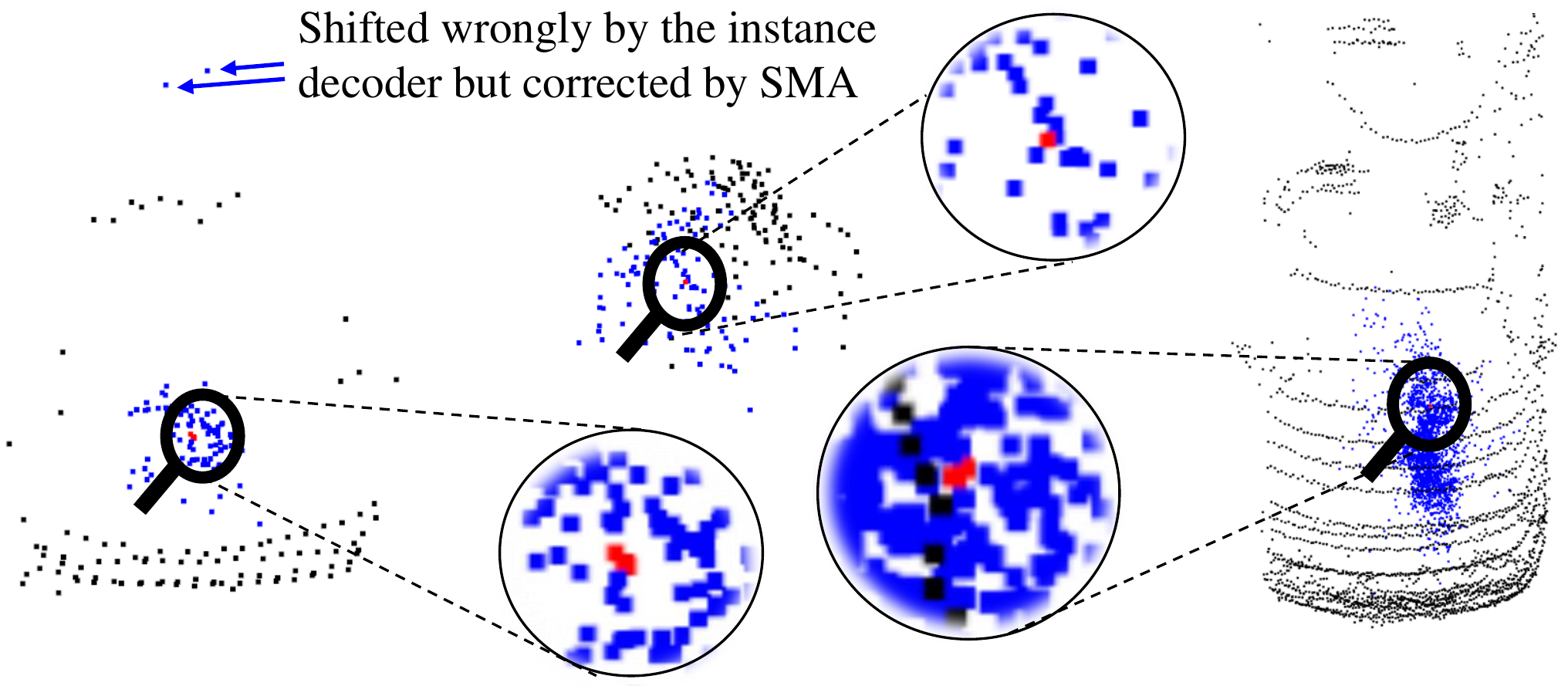}
    }
    \caption{Three examples of foreground points before and after SMAC module. The black points ($\mathbf{P}_{th}$) are from the original LiDAR sensor; \textcolor{blue}{blue} points ($\mathbf{P}_{s}$) are the positions of the points after applying the 2D offsets from the instance decoder; \textcolor{red}{red} points ($\mathbf{C}_{f}$) are the final shifted position after the proposed SMAC module.}
    \label{fig:offsets}
\end{figure}

The performance of \proposed~in comparison to other models is shown in Table \ref{bigtable} and Table \ref{bigtable_nu}. For evaluations on SemanticKITTI test dataset (Table \ref{bigtable}), the models are divided into two groups based on latency. The models in row 1-7 either do not have an available inference speed or are not capable to operate in real-time, while those in row 8-10 are known to be able to operate in real-time, with FPS's greater than 10Hz. With a PQ of $56.1\%$ and an FPS of 10.1Hz, our proposed model (row 10) obtains a good balance between run-time and accuracy. Specifically, it achieves the highest PQ for real-time models, outperforming the best real-time baseline, Panoptic-PolarNet, by $2.0\%$.

A deeper look into the other metrics reveals that SMAC-Seg is better than other real-time models in recognizing and differentiating between instances, achieving an improvement of $1.2\%$ in {RQ\textsuperscript{Th}} over Panoptic-PolarNet. This demonstrates that the addition of the SMA module to refine predictions of the instance decoder before clustering is helpful for identifying instances. In terms of metrics correlated to semantic segmentation performance, despite a reduction of $1.6\%$ in {SQ\textsuperscript{Th}} over Panoptic-PolarNet, other metrics such as SQ, {PQ\textsuperscript{St}} and mIOU see improvements of $0.6\%$, $3.6\%$ and $3.8\%$, respectively. This means that although the semantic decoder within SMAC-Seg has not obtained the best results in segmenting certain thing classes, it performs particularly well in labeling background classes. This higher mIoU and SQ can be mainly attributed to the CLSA module, which incorporates 3D geometry features that are otherwise lost after spherical projection.

For evaluations on nuScenes validation dataset (Table \ref{bigtable_nu}), because the methods for creating the instance labels for panoptic segmentation may be different across publications, we separated the models into three groups. The first (row 1-4) and second group (rows 5-6) are existing published methods grouped by inference speed, similar to Table \ref{bigtable}. The third group (row 7-10) contains the baseline models and proposed models used in our experiments. SMAC-Seg HiRes (rows 10) achieves the highest PQ out of all models, with a $15.8\%$ and $5.1\%$ improvement over the baseline models Dual Decoder (BFS) and Dual Decoder (HDBSCAN), respectively.

However, since nuScenes dataset is obtained in an urban setting and contains more objects per scene than semanticKITTI dataset, the inference time of SMAC-Seg with a high-resolution range image (row 10) is no longer real-time. By reducing the resolution from $64 \times 2048$ to $32 \times $1024, SMAC-Seg (row 9) again achieves a real-time performance with only a small trade-off in PQ.

\begin{table*}[htbp!]
{
\centering
\resizebox{1.82\columnwidth}{!}{
\begin{tabular}{l|cccc|ccc|ccc|c|c}
\hline 
Method & PQ  
& PQ\textsuperscript{$\dagger$}
& RQ
& SQ 
& PQ\textsuperscript{Th} 
& RQ\textsuperscript{Th}
& SQ\textsuperscript{Th}  
& PQ\textsuperscript{St}  
& RQ\textsuperscript{St}  
& SQ\textsuperscript{St}  
&  mIoU  
&  FPS  \\
\hline
RangeNet++ \cite{milioto2019rangenet++} + PointPillars \cite{Lang_2019_CVPR_pointpillars}
& $37.1$ & $45.9$ & $47.0$ & $75.9$ & $20.2$ & $25.2$ & $75.2$ & $49.3$ & $62.8$ & $76.5$ & $52.4$  & $2.4$\\
PanopticTrackNet \cite{hurtado2020mopt}
& $43.1$ & $50.7$ & $53.9$ & $78.8$ & $28.6$ & $35.5$ & $80.4$ & $53.6$ & $67.3$ & $77.7$ & $52.6$  & $\textbf{6.8}$ \\
KPConv \cite{thomas2019kpconv} + PointPillars \cite{Lang_2019_CVPR_pointpillars}
& $44.5$ & $52.5$ & $54.4$ & $80.0$ & $32.7$ & $38.7$ & $81.5$ & $53.1$ & $65.9$ & $79.0$ & $58.8$ & $1.9$\\
Panoster \cite{gasperini2021panoster}
& $52.7$ & $59.9$ & $64.1$ & $80.7$ & $49.4$ & $58.5$ & $83.3$ & $55.1$ & $68.2$ & $78.8$ & $59.9$  & $-$\\
DS-Net \cite{hong2020lidar}
& $55.9$ & $62.5$ & $66.7$ & $82.3$ & $55.1$ & $62.8$ & $87.2$ & $56.5$ & $69.5$ & $78.7$ & $61.6$  & $3.4^{\dagger}$\\
EfficientLPS \cite{sirohi2021efficientlps}
& $57.4$ & $63.2$ & $68.7$ & $\textbf{83.0}$ & $53.1$ & $60.5$ & $\textbf{87.8}$ & $\textbf{60.5}$ & $\textbf{74.6}$ & $\textbf{79.5}$ & $61.4$  & $-$\\
GP-S3Net \cite{razani2021gps3} 
 & $\textbf{60.0}$ & $\textbf{69.0}$ & $\textbf{72.1}$ & $82.0$ & $\textbf{65.0}$ & $\textbf{74.5}$ & $86.6$ & $56.4$ & $70.4$ & $78.7$ & $\textbf{70.8}$ & $3.7^{*}$
\\
\hline
LPSAD \cite{9340837}
& $38.0$ & $47.0$ & $48.2$ & $76.5$ & $25.6$ & $31.8$ & $76.8$ & $47.1$ & $60.1$ & $76.2$ & $50.9$   & $\textbf{11.8}$ \\
Panoptic-PolarNet \cite{Zhou2021PanopticPolarNet}
 & $54.1$ & $60.7$ & $65.0$ & $81.4$ & $\textbf{53.3}$ & $60.6$ & $\textbf{87.2}$ & $54.8$ & $68.1$ & $77.2$ & $59.5$ & $11.6$
\\
 \textbf{SMAC-Seg} [\textcolor{blue}{Ours}] 
 & $\textbf{56.1}$ & $\textbf{62.5}$ & $\textbf{67.9}$ & $\textbf{82.0}$ & $53.0$ & $\textbf{61.8}$ & $85.6$ & $\textbf{58.4}$ & $\textbf{72.3}$ & $\textbf{79.3}$ & $\textbf{63.3}$  & $10.1$
\\
\hline
\end{tabular}
}
\caption{Comparison of LiDAR panoptic segmentation performance on SemanticKITTI\cite{DBLP:conf/iccv/BehleyGMQBSG19} test dataset. Metrics are provided in [\%] and FPS is in [Hz].(*: measured with our implementation based on \cite{razani2021gps3}; $\dagger$: measured using official codebase released by the authors.)}
\label{bigtable} }
\end{table*} 

\begin{table*}[htbp!]
{
\centering
\resizebox{1.82\columnwidth}{!}{
\begin{tabular}{l|cccc|ccc|ccc|c|c}
\hline 
Method & PQ  
& PQ\textsuperscript{$\dagger$}
& RQ
& SQ 
& PQ\textsuperscript{Th} 
& RQ\textsuperscript{Th}
& SQ\textsuperscript{Th}  
& PQ\textsuperscript{St}  
& RQ\textsuperscript{St}  
& SQ\textsuperscript{St}  
&  mIoU  
&  FPS  \\
\hline
DS-Net \cite{hong2020lidar}
& $42.5$ & $51.0$ & $50.3$ & $83.6$ & $32.5$ & $38.3$ & $83.1$ & $59.2$ & $70.3$ & $\textbf{84.4}$ & $70.7$ & $-$ \\
PanopticTrackNet \cite{hurtado2020mopt}
& $50.0$ & $57.3$ & $60.6$ & $80.9$ & $45.1$ & $52.4$ & $80.3$ & $58.3$ & $74.3$ & $81.9$ & $63.1$ & $-$ \\

EfficientLPS \cite{sirohi2021efficientlps} 
& $59.2$ & $62.8$ & $70.7$ & $82.9$ & $51.8$ & $62.7$ & $80.6$ & $\textbf{71.5}$ & $\textbf{84.1}$ & $84.3$ & $69.4$  & $-$\\
GP-S3Net \cite{razani2021gps3}  
 & $\textbf{61.0}$ & $\textbf{67.5}$ & $\textbf{72.0}$ & $\textbf{84.1}$ & $\textbf{56.0}$ & $\textbf{65.2}$ & $\textbf{85.3}$ & $66.0$ & $78.7$ & $ 82.9$ & $\textbf{75.8}$ & $-$
\\
\hline
LPSAD \cite{9340837} 
& $50.4$ & $57.7$ & $62.4$ & $79.4$ & $43.2$ & $53.2$ & $80.2$ & $57.5$ & $71.7$ & $78.5$ & $62.5$ & $\textbf{22.3}^{*}$ \\
Panoptic-PolarNet \cite{Zhou2021PanopticPolarNet}
 & $\textbf{67.7}$ & $\textbf{71.0}$ & $\textbf{78.1}$ & $\textbf{86.0}$ & $\textbf{65.2}$ & $\textbf{74.0}$ & $\textbf{87.2}$ & $\textbf{71.9}$ & $\textbf{84.9}$ & $\textbf{83.9}$ & $\textbf{69.3}$ & $10.1$
\\
\hline
Dual-Dec UNet w/ BFS [Our Baseline]  
 & $52.6$ & $58.0$ & $64.3$ & $79.7$ & $39.2$ & $49.0$ & $77.5$ & $66.0$ & $79.5$ & $81.9$ & $71.1$ & $\textbf{20.1}$ \\
Dual-Dec UNet w/ HDBSCAN [Our Baseline]
 & $63.3$ & $68.8$ & $75.2$ & $83.5$ & $60.8$ & $70.9$ & $85.0$ & $66.0$ & $79.5$ & $81.9$ & $71.1$ & $18.8$
\\
\textbf{SMAC-Seg}  [\textcolor{blue}{Ours}]
 & $67.0$ & $71.8$ & $78.2$ & $85.0$ & $65.2$ & $74.2$ & $87.1$ & $\textbf{68.8}$ & $\textbf{82.2}$ & $82.9$ & $\textbf{72.2}$ & $14.5$
 \\
 \textbf{SMAC-Seg HiRes}  [\textcolor{blue}{Ours}]
 & $\textbf{68.4}$ & $\textbf{73.4}$ & $\textbf{79.7}$ & $\textbf{85.2}$ & $\textbf{68.0}$ & $\textbf{77.2}$ & $\textbf{87.3}$ & $\textbf{68.8}$ & $82.1$ & $\textbf{83.0}$ & $71.2$ & $6.2$
\\
\hline
\end{tabular}
}
\caption{Comparison of LiDAR panoptic segmentation performance on nuScenes \cite{caesar2020nuscenes} validation dataset. Metrics are provided in [\%] and FPS is in [Hz].(*: measured with our implementation based on \cite{9340837})}
\label{bigtable_nu}
}
\end{table*} 
\subsection{Qualitative Evaluation}
\label{sec:quali}

In Fig. \ref{fig:qualitative}, we demonstrate the panoptic segmentation performance of the SMAC-Seg model by comparing its inference results to LPSAD, our implementation based on \cite{9340837}. For a relatively simple scene from SemanticKITTI dataset with a row of cars, SMAC-Seg segments the instance points without errors, while LPSAD identifies the car on the right as two separate instances. In the example from nuScenes, the scene is more complex, with an intersection and a mix of cars, trucks, and cyclists present. With few sparse points describing each instance and variations in object size, correctly predicting object centroids and segmenting instances is more difficult. In cases like this, LPSAD is prone to errors. The large truck on the left of the frame is segmented into two instances, while the cyclist immediately to its right is also segmented into two. At the bottom of the frames, as two cars are close to each other, LPSAD mistakenly assigns half of the car on the right to be part of the car on the left. In comparison, SMAC-Seg is able to segment each instance in a complex scene with high accuracy. The ability for SMAC-Seg to accurately group points within the same instance can be primarily attributed to the directional guidance during offset correction within the SMA module and the supervision of attract and loss. Meanwhile, the strength of SMAC-Seg in accurately differentiating between two closely positioned objects is made possible by the supervision of the centroid-aware repel loss function.

From Fig. \ref{fig:offsets}, we further present the effectiveness of the SMA module. The blue points are the predicted centroids after shifting each foreground point by the regressed offsets from instance decoder. Although most points are shifted towards the center of the instance, some indicated in the figure are shifted to a wrong direction. A clustering algorithm applied here is not ideal, as the object will be over-segmented. This indicates that the supervision by L2-loss within instance decoder is insufficient to aid clustering. On the other hand, the addition of the SMA module before BFS clustering effectively corrects the offset predictions and aggregates each object cluster towards its centroid. This additional relocation of points allows the clustering algorithm to return a more accurate result, as shown by the final shifted points in red.

\subsection{Ablation Studies}
To further demonstrate the influence of each proposed component, we conducted ablation studies on the validation set of SemanticKITTI. 
We first compare our proposed SMAC module against various clustering algorithms (row 1-5 of Table \ref{tab:Ablation}), including BFS, HDBSCAN \cite{campello_hdbscan}, Meanshift \cite{comaniciu_meanshift}, and DS \cite{hong2020lidar}. The results show that HDBSCAN and Meanshift clustering achieve a slightly better PQ when segmenting the shifted foreground points, while DS and SMAC with the baseline backbone are the only candidates here that are capable of operating in real-time. When benchmarking against DS, our proposed SMAC runs $30 ms$ faster with a PQ improvement of $0.5\%$. Note that SMAC module is a combination of SMA and BFS clustering, when comparing with the BFS baseline (row 1), the proposed module not only increases PQ from $44.9\%$ to $51.9\%$ but decreases the runtime spent on clustering significantly as the foreground points are located together with other neighbouring points that belong to the same instance and away from others (as seen in Fig.\ref{fig:offsets}). In addition, from row 6-10, we show the contribution of each proposed components, namely Cross Block, CLSA, Centroid-aware Repel Loss, and the combinations of them. We also report the FPS of each entry and bold the ones that operate in real-time.

\begin{table}[h]
\begin{center}
\scalebox{0.65}
{
\begin{tabular}{ c|ccc|ccccc|c|c  }
\hline 
\multicolumn{1}{c|}{\textbf{Architecture}} &
\multicolumn{1}{c}{\textbf{\begin{turn}{45} crossBlock \end{turn} }} &
\multicolumn{1}{c}{\textbf{\begin{turn}{45} CLSA \end{turn} }} &
\multicolumn{1}{c}{\textbf{\begin{turn}{45} Repel \end{turn} }} &

\multicolumn{1}{|c}{\textbf{\begin{turn}{45} BFS \end{turn} }} &
\multicolumn{1}{c}{\textbf{\begin{turn}{45} HDB \end{turn} }} &
\multicolumn{1}{c}{\textbf{\begin{turn}{45} MS \end{turn} }} &
\multicolumn{1}{c}{\textbf{\begin{turn}{45} DS \end{turn} }} &
\multicolumn{1}{c}{\textbf{\begin{turn}{45} SMAC \end{turn} }} &
\multicolumn{1}{|c}{\textbf{mPQ}} &
\multicolumn{1}{|c}{\textbf{FPS}} \\

 \hline \hline 
\multirow{4}{*}{Baseline}
&     &   &       &\checkmark & & & &   \multicolumn{1}{c|}{ } & $44.9$ & $8.6$ \rule{0pt}{3ex}\\

&     &   &       & &\checkmark & & &   \multicolumn{1}{c|}{ } & $52.7$ & $4.8$ \rule{0pt}{3ex}\\

&     &   &       & & &\checkmark & &   \multicolumn{1}{c|}{ } & $52.8$ & $3.2$ \rule{0pt}{3ex}\\

&     &   &       & & & &\checkmark &   \multicolumn{1}{c|}{ } & $51.4$ & $\textbf{11.1}$ \rule{0pt}{3ex}\\

\hline
\multirow{6}{*}{Proposed}
&     &   &       & & & & &   \checkmark & $51.9$ & $\textbf{16.1}$ \rule{0pt}{3ex}\\

& \checkmark &   & &\checkmark      & & &    & \multicolumn{1}{c|}{ }  & $46.2$ & $7.7$ \rule{0pt}{3ex}\\ 

& \checkmark & \checkmark &      &\checkmark      & & &   &  \multicolumn{1}{c|}{ }  & $48.6$ & $6.6$ \rule{0pt}{3ex}\\ 

& \checkmark & \checkmark &   &         & & &   & \checkmark  & $54.1$ & $\textbf{10.1}$ \rule{0pt}{3ex}\\ 

& \checkmark &  & \checkmark  & & &  &  & \checkmark  & $52.9$ & $\textbf{15.5}$ \rule{0pt}{3ex}\\ 

& \checkmark & \checkmark &\checkmark      &   & &  &  & \checkmark  & $\textbf{55.8}$ & $\textbf{10.1}$  \rule{0pt}{3ex}\\ \hline

\end{tabular}}
\end{center}
\vspace{-10px}
\caption{Ablation study of the proposed components vs baseline.}
\label{tab:Ablation}
\end{table}

\begin{figure}[htb!]
    \centering
    \includegraphics[width=\linewidth]{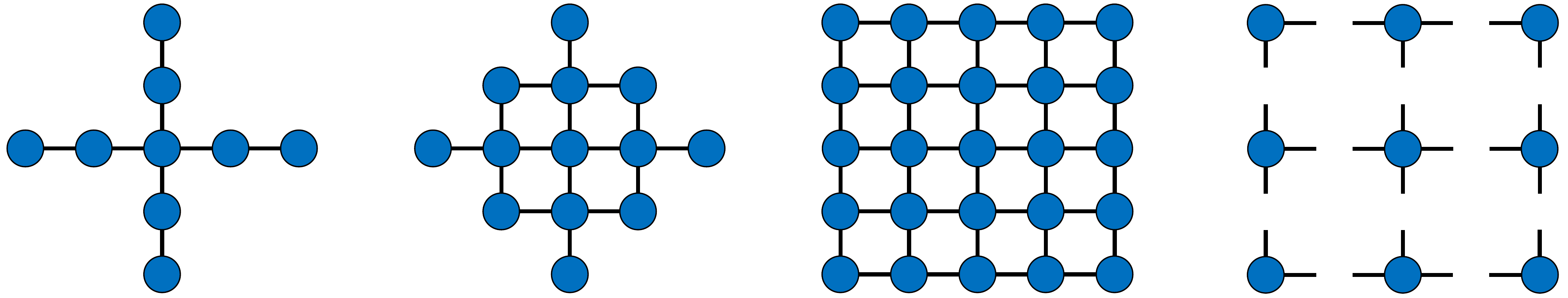}
    \caption{Examples of kernel size of 5. From left to right: cross kernel, diamond kernel, square dense kernel, and square kernel with dilation.}
    \label{fig:kernel}
\end{figure}

In addition, We evaluated the effectiveness of convolutions with CLSA on the semantic segmentation tasks with various kernel shapes as illustrated in Fig. \ref{fig:kernel}. From the results in Table \ref{tab:kernel}, cross kernel is the most effective in utilizing the local neighbourhood geometry. Using CLSA with square dense kernels performs relatively worse than others; with large square kernels, the network tends to memorize some of the local geometry and introduces overfitting to the training set. Nevertheless, CLSA is highly effective as it improves the mean IoU  by $3.3\%$ comparing to the baseline. It learns to recover the local 3D geometry features, which range-based (RV) approaches lose after the spherical projection.

\label{sec:ablation}
\begin{table}[htb!]
\begin{center}
\scalebox{0.89}
{
\begin{tabular}{ c|c|c|c  }
\hline 
\multicolumn{1}{c|}{Architecture} &
\multicolumn{1}{c|}{Sampling Region} &
\multicolumn{1}{c|}{mIoU} &
\multicolumn{1}{c}{Improvement} \\

 \hline \hline 
 \multirow{1}{*}{baseline}
& $-$ & $58.9$ & $-$  \rule{0pt}{3ex}\\  \hline
 
 \multirow{4}{*}{CLSA}
&Cross Kernel & $\textbf{62.2}$ & $\textbf{+3.3}$  \rule{0pt}{3ex}\\ \cline{2-4}
&Diamond Kernel & $61.7$ & $+2.8$  \rule{0pt}{3ex}\\ \cline{2-4}
&Square Dense Kernel & $61.1$ & $+2.2$  \rule{0pt}{3ex}\\ \cline{2-4}
&Square Kernel with Dilation & $61.7$ & $+2.8$  \rule{0pt}{3ex}\\ \hline

\end{tabular}
}
\end{center}
\caption{Ablation study of using different neighbourhood sampling with local spatial attention.}
\label{tab:diamond}
\end{table}

Lastly, we conducted experiments on the SMAC module to see how various BEV grid sizes and directional kernel sizes could affect the performance. In particular, we chose to run a total of six experiments with grid size of $\{0.3, 0.5, 1.0\}$m, and with kernel size of $3$ and $7$ for each. From Tab.\ref{tab:kernel}, the combination of grid size $0.5$ and kernel size $7$ achieves the best performance.

\begin{table}[htb!]
\begin{center}
\scalebox{0.99}
{
\begin{tabular}{ c|c|cccc  }
\hline 
\multicolumn{1}{c|}{Grid Size} &
\multicolumn{1}{c|}{Kernel Size} &
\multicolumn{1}{c}{PQ} &
\multicolumn{1}{c}{PQ\textsuperscript{Th}} &
\multicolumn{1}{c}{RQ\textsuperscript{Th}} &
\multicolumn{1}{c}{SQ\textsuperscript{Th}} \\

 \hline \hline 
 \multirow{2}{*}{0.3}
&$3$ & $55.5$ & $56.2$ & $63.7$      &    $\textbf{76.6}$   \rule{0pt}{3ex}\\ \cline{2-6}
&$7$ & $55.5$ & $57.1$ & $64.9$      &    $76.4$   \rule{0pt}{3ex}\\ \hline
 
 \multirow{2}{*}{0.5}
&$3$ & $55.3$ & $56.2$ & $64.3$      &    $76.0$   \rule{0pt}{3ex}\\ \cline{2-6}

&$7$ & $\textbf{55.8}$ & $\textbf{57.4}$ & $\textbf{65.5}$      &    $76.4$  \rule{0pt}{3ex}\\ \hline

\multirow{2}{*}{1.0}
&$3$ & $55.0$ & $55.6$ & $63.8$      &    $75.8$   \rule{0pt}{3ex}\\ \cline{2-6}

&$7$ & $55.1$ & $55.4$ & $63.4$      &    $76.1$   \rule{0pt}{3ex}\\ \hline


\end{tabular}
}
\end{center}
\caption{Ablation study of using different grid size and kernel size in SMAC.}
\label{tab:kernel}
\end{table}

\section{CONCLUSION}
We have proposed SMAC-Seg, a novel end-to-end real-time panoptic segmentation network for LiDAR point cloud. In particular, the model comprises of a robust encoder including a novel CLSA block to recover 3D geometric features lost due to projection and multiple Cross blocks to learn and fuse multi-scale features. To segment each object, each foreground cluster is learned to effectively aggregate towards its centroid in the proposed SMAC module. Further, a novel centroid-aware repel loss is added to supervise the model in differentiating each cluster with its neighbour. From the evaluation on both large-scale datasets of SemanticKITTI and nuScenes, we have shown that SMAC-Seg achieves state-of-the-art performance among all published real-time methods at the time of submission. 


\bibliographystyle{IEEEtran}
\bibliography{bib}

\begin{thebibliography}{10}
\providecommand{\url}[1]{#1}
\csname url@rmstyle\endcsname
\providecommand{\newblock}{\relax}
\providecommand{\bibinfo}[2]{#2}
\providecommand\BIBentrySTDinterwordspacing{\spaceskip=0pt\relax}
\providecommand\BIBentryALTinterwordstretchfactor{4}
\providecommand\BIBentryALTinterwordspacing{\spaceskip=\fontdimen2\font plus
\BIBentryALTinterwordstretchfactor\fontdimen3\font minus
  \fontdimen4\font\relax}
\providecommand\BIBforeignlanguage[2]{{%
\expandafter\ifx\csname l@#1\endcsname\relax
\typeout{** WARNING: IEEEtran.bst: No hyphenation pattern has been}%
\typeout{** loaded for the language `#1'. Using the pattern for}%
\typeout{** the default language instead.}%
\else
\language=\csname l@#1\endcsname
\fi
#2}}

\bibitem{mohan2021ijcv}
R.~Mohan and A.~Valada, ``Efficientps: Efficient panoptic segmentation,''
  \emph{International Journal of Computer Vision}, vol. 129, pp. 1551 -- 1579,
  2020.

\bibitem{Cheng2020panopticdeeplab}
B.~Cheng, M.~Collins, Y.~Zhu, T.~Liu, H.~Adam, and L.-C. Chen,
  ``Panoptic-deeplab: A simple, strong, and fast baseline for bottom-up
  panoptic segmentation,'' 06 2020, pp. 12\,472--12\,482.

\bibitem{9340837}
A.~{Milioto}, J.~{Behley}, C.~{McCool}, and C.~{Stachniss}, ``Lidar panoptic
  segmentation for autonomous driving,'' in \emph{2020 IEEE/RSJ International
  Conference on Intelligent Robots and Systems (IROS)}, 2020, pp. 8505--8512.

\bibitem{razani2021gps3}
R.~Razani, R.~Cheng, E.~Li, E.~Taghavi, Y.~Ren, and L.~Bingbing, ``Gp-s3net:
  Graph-based panoptic sparse semantic segmentation network,'' \emph{arXiv
  preprint arXiv:2108.08401}, 2021.

\bibitem{qi2017pointnet}
C.~R. Qi, H.~Su, K.~Mo, and L.~J. Guibas, ``Pointnet: Deep learning on point
  sets for 3d classification and segmentation,'' in \emph{Proceedings of the
  IEEE conference on computer vision and pattern recognition}, 2017, pp.
  652--660.

\bibitem{thomas2019kpconv}
H.~Thomas, C.~R. Qi, J.-E. Deschaud, B.~Marcotegui, F.~Goulette, and L.~J.
  Guibas, ``Kpconv: Flexible and deformable convolution for point clouds,'' in
  \emph{Proceedings of the IEEE International Conference on Computer Vision},
  2019, pp. 6411--6420.

\bibitem{cheng2021af2s3net}
R.~Cheng, R.~Razani, E.~Taghavi, E.~Li, and B.~Liu, ``(af)2-s3net: Attentive
  feature fusion with adaptive feature selection for sparse semantic
  segmentation network,'' in \emph{Proceedings of the IEEE/CVF Conference on
  Computer Vision and Pattern Recognition (CVPR)}, June 2021.

\bibitem{zhou2020cylinder3d}
X.~Zhu, H.~Zhou, T.~Wang, F.~Hong, Y.~Ma, W.~Li, H.~Li, and D.~Lin,
  ``Cylindrical and asymmetrical 3d convolution networks for lidar
  segmentation,'' in \emph{Proceedings of the IEEE/CVF Conference on Computer
  Vision and Pattern Recognition (CVPR)}, June 2021, pp. 9939--9948.

\bibitem{cortinhal2020salsanext}
T.~Cortinhal, G.~Tzelepis, and E.~E. Aksoy, ``Salsanext: Fast semantic
  segmentation of lidar point clouds for autonomous driving,'' in \emph{2020
  IEEE Intelligent Vehicles Symposium (IV)}, 2020, pp. 655--661.

\bibitem{milioto2019rangenet++}
A.~Milioto, I.~Vizzo, J.~Behley, and C.~Stachniss, ``Rangenet++: Fast and
  accurate lidar semantic segmentation,'' in \emph{Proc. of the IEEE/RSJ Intl.
  Conf. on Intelligent Robots and Systems (IROS)}, 2019.

\bibitem{razani2021litehdseg}
R.~Razani, R.~Cheng, E.~Taghavi, and L.~Bingbing, ``Lite-hdseg: Lidar semantic
  segmentation using lite harmonic dense convolutions,'' in \emph{2021 IEEE
  International Conference on Robotics and Automation (ICRA)}.\hskip 1em plus
  0.5em minus 0.4em\relax IEEE, 2021.

\bibitem{zhang2020polarnet}
Y.~Zhang, Z.~Zhou, P.~David, X.~Yue, Z.~Xi, B.~Gong, and H.~Foroosh,
  ``Polarnet: An improved grid representation for online lidar point clouds
  semantic segmentation,'' in \emph{Proceedings of the IEEE/CVF Conference on
  Computer Vision and Pattern Recognition}, 2020, pp. 9601--9610.

\bibitem{10.1007/978-3-030-11009-3_11}
M.~Simon, S.~Milz, K.~Amende, and H.-M. Gross, ``Complex-yolo: An
  euler-region-proposal for real-time 3d object detection on point clouds,'' in
  \emph{Computer Vision -- ECCV 2018 Workshops}, L.~Leal-Taix{\'e} and S.~Roth,
  Eds.\hskip 1em plus 0.5em minus 0.4em\relax Cham: Springer International
  Publishing, 2019, pp. 197--209.

\bibitem{Zhou2021PanopticPolarNet}
Z.~Zhou, Y.~Zhang, and H.~Foroosh, ``Panoptic-polarnet: Proposal-free lidar
  point cloud panoptic segmentation,'' in \emph{Proceedings of the IEEE/CVF
  Conference on Computer Vision and Pattern Recognition (CVPR)}, 2021.

\bibitem{hong2020lidar}
F.~Hong, H.~Zhou, X.~Zhu, H.~Li, and Z.~Liu, ``Lidar-based panoptic
  segmentation via dynamic shifting network,'' in \emph{Proceedings of the
  IEEE/CVF Conference on Computer Vision and Pattern Recognition (CVPR)}, June
  2021.

\bibitem{DBLP:conf/iccv/BehleyGMQBSG19}
\BIBentryALTinterwordspacing
J.~Behley, M.~Garbade, A.~Milioto, J.~Quenzel, S.~Behnke, C.~Stachniss, and
  J.~Gall, ``Semantickitti: {A} dataset for semantic scene understanding of
  lidar sequences,'' in \emph{2019 {IEEE/CVF} International Conference on
  Computer Vision, {ICCV} 2019, Seoul, Korea (South), October 27 - November 2,
  2019}.\hskip 1em plus 0.5em minus 0.4em\relax {IEEE}, 2019, pp. 9296--9306.
  [Online]. Available: \url{https://doi.org/10.1109/ICCV.2019.00939}
\BIBentrySTDinterwordspacing

\bibitem{caesar2020nuscenes}
H.~Caesar, V.~Bankiti, A.~H. Lang, S.~Vora, V.~E. Liong, Q.~Xu, A.~Krishnan,
  Y.~Pan, G.~Baldan, and O.~Beijbom, ``nuscenes: A multimodal dataset for
  autonomous driving,'' in \emph{Proceedings of the IEEE/CVF Conference on
  Computer Vision and Pattern Recognition}, 2020, pp. 11\,621--11\,631.

\bibitem{panopticMetric}
A.~Kirillov, K.~He, R.~Girshick, C.~Rother, and P.~Dollar, ``Panoptic
  segmentation,'' \emph{2019 IEEE/CVF Conference on Computer Vision and Pattern
  Recognition (CVPR)}, 2019.

\bibitem{zhao2017pspnet}
H.~Zhao, J.~Shi, X.~Qi, X.~Wang, and J.~Jia, ``Pyramid scene parsing network,''
  07 2017, pp. 6230--6239.

\bibitem{he2017maskrcnn}
K.~He, G.~Gkioxari, P.~Dollar, and R.~Girshick, ``Mask r-cnn,'' 10 2017, pp.
  2980--2988.

\bibitem{li2019attention}
Y.~Li, X.~Chen, Z.~Zheng, L.~Xie, G.~Huang, D.~Du, and X.~Wang,
  ``Attention-guided unified network for panoptic segmentation,'' 06 2019, pp.
  7019--7028.

\bibitem{DBLP:journals/corr/abs-1902-05093}
\BIBentryALTinterwordspacing
T.~Yang, M.~D. Collins, Y.~Zhu, J.~Hwang, T.~Liu, X.~Zhang, V.~Sze,
  G.~Papandreou, and L.~Chen, ``Deeperlab: Single-shot image parser,''
  \emph{CoRR}, vol. abs/1902.05093, 2019. [Online]. Available:
  \url{http://arxiv.org/abs/1902.05093}
\BIBentrySTDinterwordspacing

\bibitem{gao2020ssap}
N.~Gao, Y.~Shan, Y.~Wang, X.~Zhao, and K.~Huang, ``Ssap: Single-shot instance
  segmentation with affinity pyramid,'' \emph{IEEE Transactions on Circuits and
  Systems for Video Technology}, vol.~PP, pp. 1--1, 04 2020.

\bibitem{wu2020bidirectional}
Y.~Wu, Z.~Gengwei, Y.~Gao, X.~Deng, K.~Gong, X.~Liang, and L.~Lin,
  ``Bidirectional graph reasoning network for panoptic segmentation,'' 06 2020,
  pp. 9077--9086.

\bibitem{cheng2021s3net}
R.~Cheng, R.~Razani, Y.~Ren, and B.~Liu, ``S3net: 3d lidar sparse semantic
  segmentation network,'' in \emph{2021 IEEE International Conference on
  Robotics and Automation (ICRA)}.\hskip 1em plus 0.5em minus 0.4em\relax IEEE,
  2021.

\bibitem{tang2020searching}
H.~Tang, Z.~Liu, S.~Zhao, Y.~Lin, J.~Lin, H.~Wang, and S.~Han, ``Searching
  efficient 3d architectures with sparse point-voxel convolution,'' in
  \emph{European Conference on Computer Vision}, 2020.

\bibitem{qi2017pointnet++}
C.~R. Qi, L.~Yi, H.~Su, and L.~J. Guibas, ``Pointnet++: Deep hierarchical
  feature learning on point sets in a metric space,'' in \emph{Advances in
  neural information processing systems}, 2017, pp. 5099--5108.

\bibitem{hu2020randla}
Q.~Hu, B.~Yang, L.~Xie, S.~Rosa, Y.~Guo, Z.~Wang, N.~Trigoni, and A.~Markham,
  ``Randla-net: Efficient semantic segmentation of large-scale point clouds,''
  in \emph{Proceedings of the IEEE/CVF Conference on Computer Vision and
  Pattern Recognition}, 2020, pp. 11\,108--11\,117.

\bibitem{gerdzhev2021tornadonet}
M.~Gerdzhev, R.~Razani, E.~Taghavi, and B.~Liu, ``Tornado-net: multiview total
  variation semantic segmentation with diamond inception module,'' in
  \emph{2021 IEEE International Conference on Robotics and Automation
  (ICRA)}.\hskip 1em plus 0.5em minus 0.4em\relax IEEE, 2021.

\bibitem{DBLP:journals/corr/abs-2003-02371}
\BIBentryALTinterwordspacing
J.~Behley, A.~Milioto, and C.~Stachniss, ``A benchmark for lidar-based panoptic
  segmentation based on {KITTI},'' \emph{CoRR}, vol. abs/2003.02371, 2020.
  [Online]. Available: \url{https://arxiv.org/abs/2003.02371}
\BIBentrySTDinterwordspacing

\bibitem{Lang_2019_CVPR_pointpillars}
A.~H. Lang, S.~Vora, H.~Caesar, L.~Zhou, J.~Yang, and O.~Beijbom,
  ``Pointpillars: Fast encoders for object detection from point clouds,'' in
  \emph{Proceedings of the IEEE/CVF Conference on Computer Vision and Pattern
  Recognition (CVPR)}, June 2019.

\bibitem{hurtado2020mopt}
J.~V. Hurtado, R.~Mohan, W.~Burgard, and A.~Valada, ``Mopt: Multi-object
  panoptic tracking,'' \emph{The IEEE Conference on Computer Vision and Pattern
  Recognition (CVPR) Workshop on Scalability in Autonomous Driving}, 2020.

\bibitem{sirohi2021efficientlps}
K.~Sirohi, R.~Mohan, D.~Büscher, W.~Burgard, and A.~Valada, ``Efficientlps:
  Efficient lidar panoptic segmentation,'' 2021.

\bibitem{gasperini2021panoster}
S.~Gasperini, M.-A. Mahani, A.~Marcos-Ramiro, N.~Navab, and F.~Tombari,
  ``Panoster: End-to-end panoptic segmentation of lidar point clouds,''
  \emph{IEEE Robotics and Automation Letters}, vol.~PP, pp. 1--1, 02 2021.

\bibitem{campello_hdbscan}
R.~J. G.~B. Campello, D.~Moulavi, and J.~Sander, ``Density-based clustering
  based on hierarchical density estimates,'' \emph{Advances in Knowledge
  Discovery and Data Mining Lecture Notes in Computer Science}, p. 160–172,
  2013.

\bibitem{comaniciu_meanshift}
D.~Comaniciu and P.~Meer, ``Mean shift: a robust approach toward feature space
  analysis,'' \emph{IEEE Transactions on Pattern Analysis and Machine
  Intelligence}, vol.~24, no.~5, p. 603–619, 2002.

\end{thebibliography}

\end{document}